# Performance Evaluation of Machine Learning-based Algorithm and Taguchi Algorithm for the Determination of the Hardness Value of the Friction Stir Welded AA 6262 Joints at a Nugget Zone

Akshansh Mishra[1*], Eyob Messele Sefene[2], Gopikrishna Nidigonda[3], Assefa Asmare Tsegaw[2]

[1] Department of Chemistry, Materials, and Chemical Engineering "Giulio Natta", Politecnico di Milano, Milan, Italy
[2,4] Bahir Dar Institute of Technology, Faculty of Mechanical and Industrial Engineering, Bahir Dar, 6000, Ethiopia
[3] Department of Mechanical Engineering, Sree Chaitanya College of Engineering, Karimnagar, Telangana. India

[1]**Orcid Id:** https://orcid.org/0000-0003-4939-359X

[1]**Email id:** akshansh.mishra@mail.polimi.it

## Abstract

Nowadays, industry 4.0 plays a tremendous role in the manufacturing industries for increasing the amount of data and accuracy in modern manufacturing systems. Thanks to artificial intelligence, particularly machine learning, big data analytics have dramatically amended, and manufacturers easily exploit organized and unorganized data. This study utilized hybrid optimization algorithms to find friction stir welding and optimal hardness value at the nugget zone. A similar AA 6262 material was used and welded in a butt joint configuration. Tool rotational speed (RPM), tool traverse speed (mm/min), and the plane depth (mm) are used as controllable parameters and optimized using Taguchi L9, Random Forest, and XG Boost machine learning tools. Analysis of variance was also conducted at a 95% confidence interval for identifying the significant parameters. The result indicated that the coefficient of determination from Taguchi L9 orthogonal array is 0.91 obtained while Random Forest and XG Boost algorithm imparted 0.62 and 0.65, respectively.

**Keywords:** Friction Stir Welding; Taguchi; Machine Learning; Hardness; Nugget Zone

## 1. Introduction

FSW (friction stir welding) is a solid-state joining process that uses frictional heat generated by a rotating tool to fuse materials [1-2]. Using a profiled probe and shoulder, the non-consumable tool is turned and plunged into the interface between two work components. It then heats and softens the material as it goes down the joint line. The shoulder also contains this plasticized material, which is physically joined to produce a solid phase weld [3-6]. This method is commonly used in industry to mix all cast, rolled, and extruded aluminum alloys [7-11]. Depending on the alloy quality and capacity of the FSW equipment, FSW has been shown to weld aluminum alloy butt joints with a thickness of between 0.3mm and 75mm in a single pass. Magnesium, titanium, copper, nickel, and steel alloys, as well as polymers and metal matrix composites, have all been employed using FSW (MMC). [12-20]. This approach can also be used to combine materials that are incompatible, such as aluminum and steel. FSW has worked with a variety of industries, including aircraft, shipbuilding, rail, electronics, and electric vehicle battery trays. From the introduction of friction stir welding in 1990 until the present, 88 researchers have attempted to optimize friction stir welding process parameters using various methodologies, according to Scopus data. They used a statistical tool called the Taguchi method, which was central in the reviewed literature and presented a relatively large number of co-occurrences with practically all other statistical analysis methods, as shown in Fig 1. and Fig 2 shows the Co-occurrence map research in the Friction Stir Welding Process, which used Machine Learning algorithms.

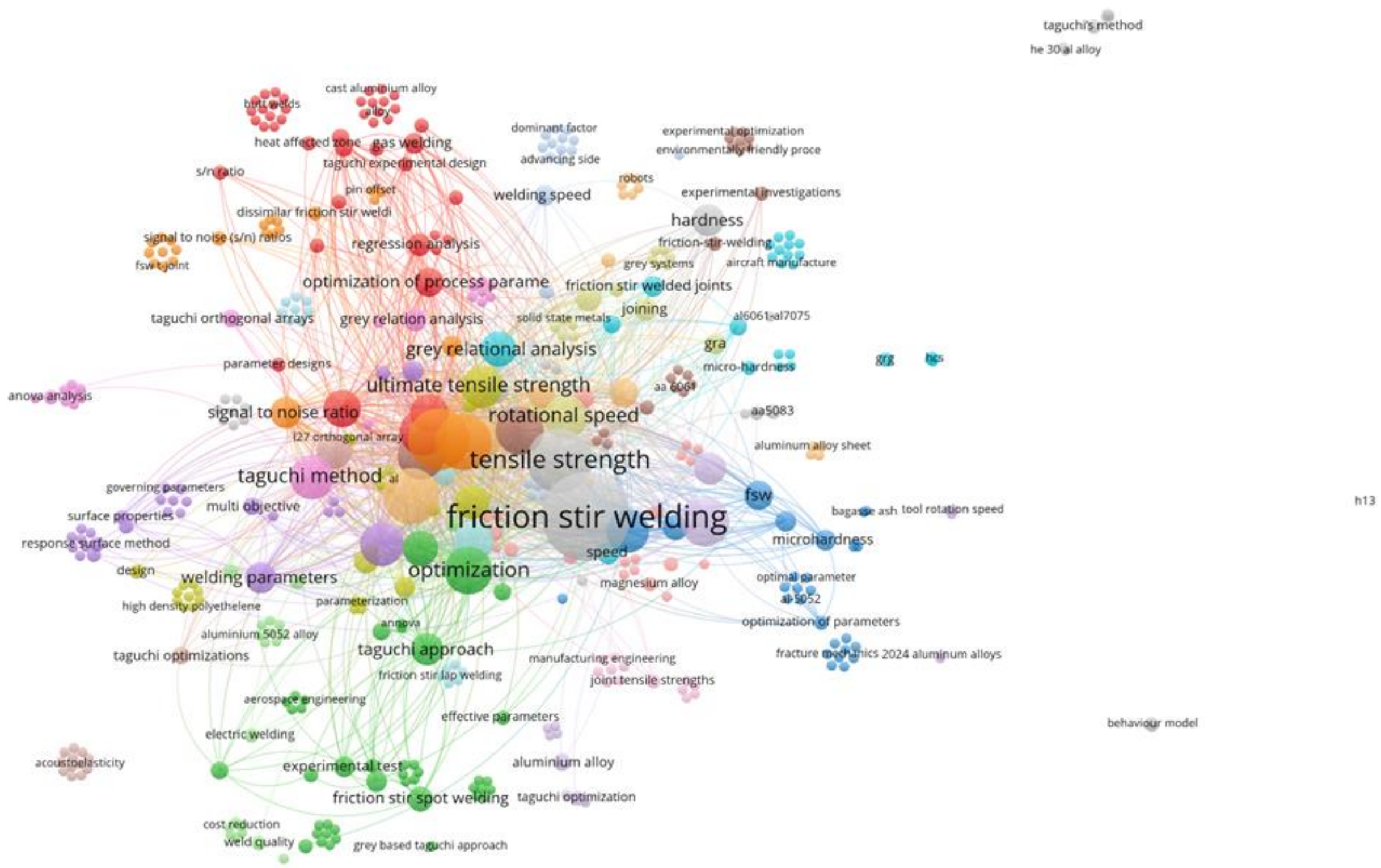

**Fig 1.** Co-occurrence analysis map of the application of the Taguchi algorithm in the Friction Stir Welding Process

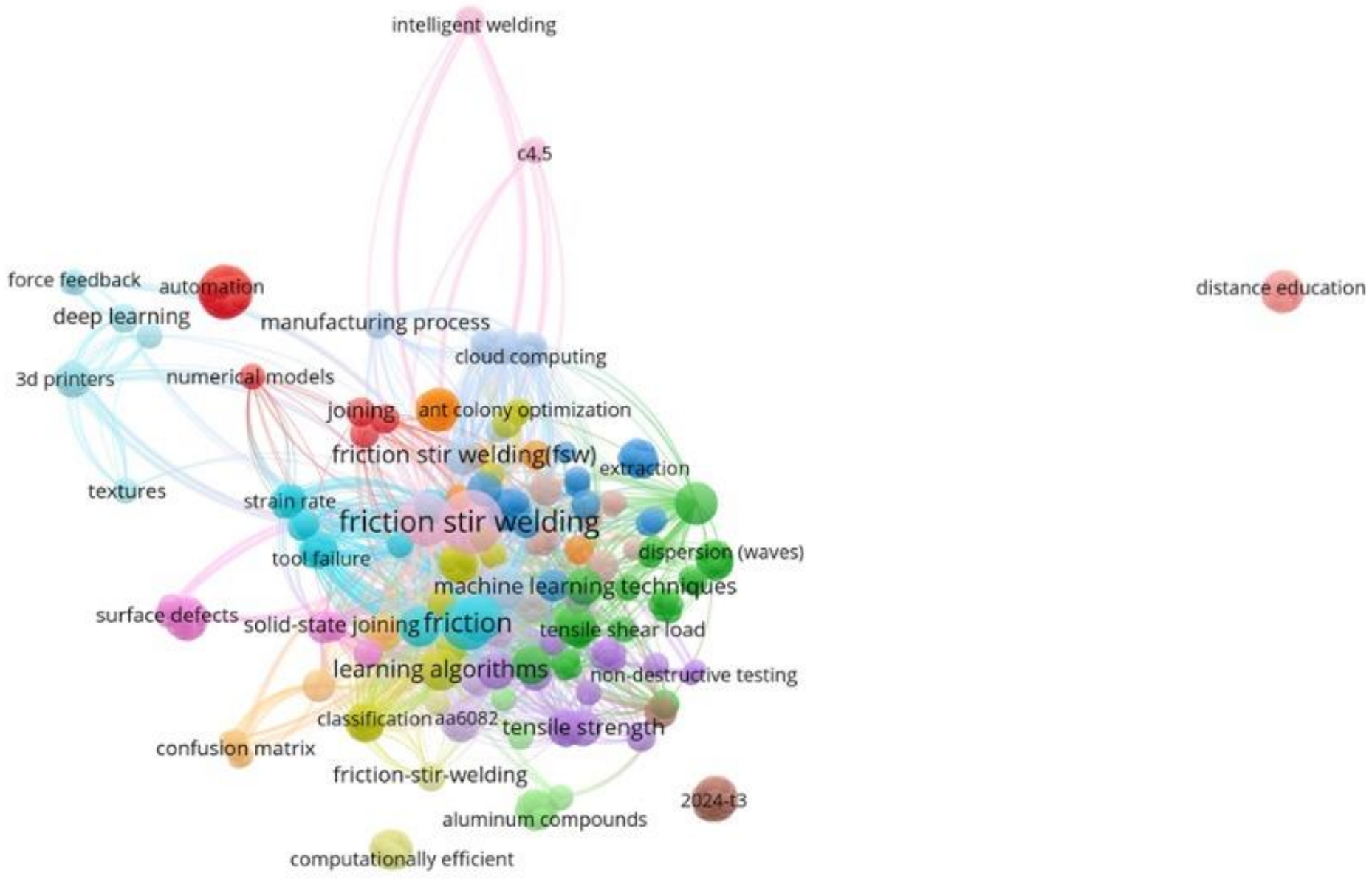

**Fig 2.** Co-occurrence analysis map of the application of Machine Learning algorithms in the Friction Stir Welding Process

## 2. Experimental Procedure

Three process parameters were used for this study: Tool Rotational Speed (RPM), Tool Travel Speed (mm/min), and Axial Force (KN). The settings were constantly updated and created various joints utilizing the RV Machine Tool's FSW machine. The test specimens are made of aluminum alloy 6262 rough plates 100 mm long, 50 mm wide, and 6 mm thick before and after welding. Fig 3 shows the tool design used in the experimental work.

Table 1 shows the chemical composition of AA 6262. Table 2 indicates the input parameters and output parameters in the present study.

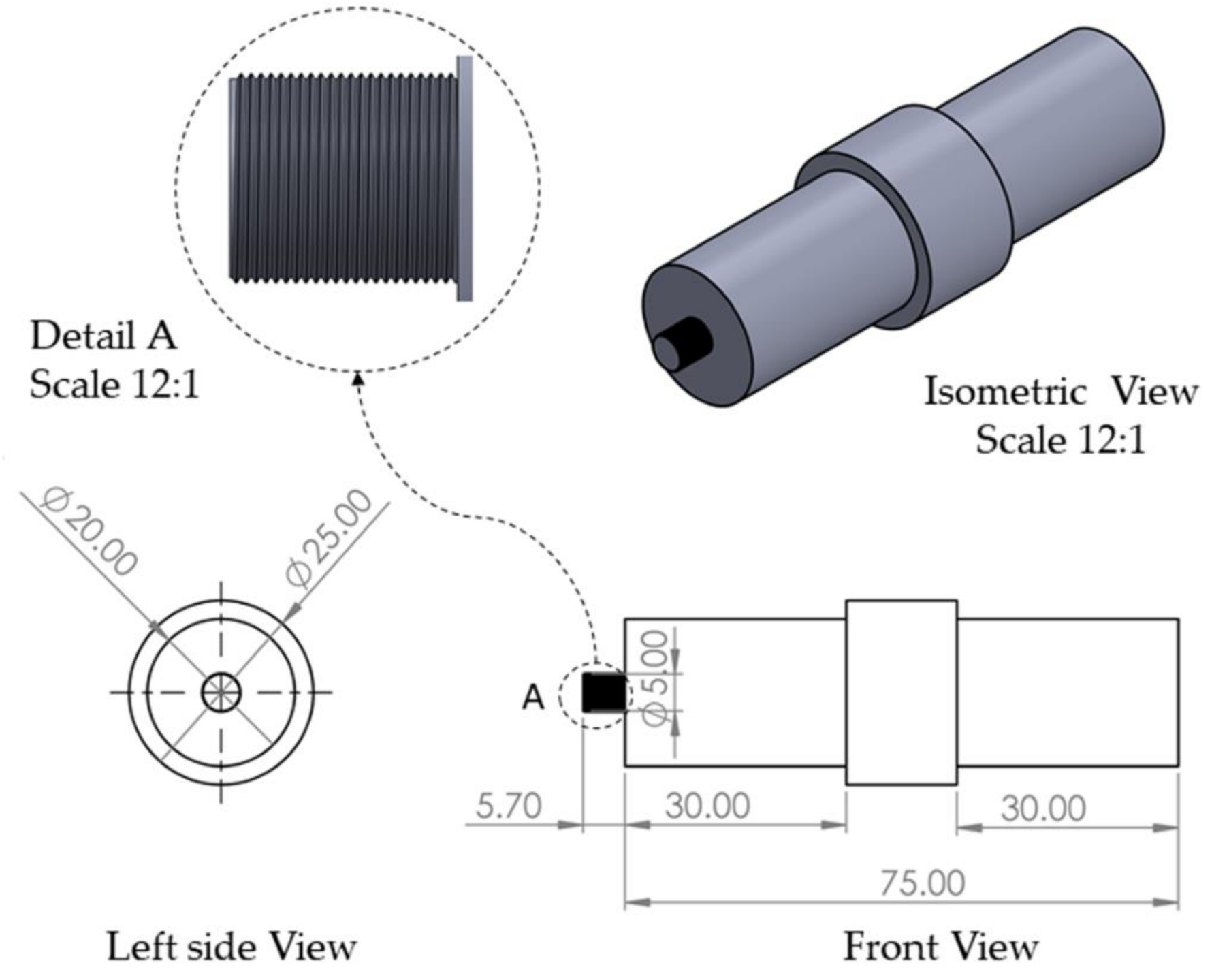

**Fig 3.** Two-dimensional Friction Stir Welding Tool Design

**Table 1.** Chemical Composition of Aluminium alloy 6262

| Element | Si | Fe | Cu | Cr | Mn | Mg | Zn |
|---|---|---|---|---|---|---|---|
| **% Present** | 0.4-0.8 | 0.0-0.7 | 0.4-1.4 | 0.0-0.2 | 0.0-0.15 | 0.8-1.2 | 0.0-0.25 |

**Table 2.** Input parameters used for preparing the given Friction Stir Welded samples

| S.No | Sample ID | Tool rotational speed (rpm) | Tool traverse speed (mm/min) | Plan depth (mm) | Hardness at Nugget Zone |
|---|---|---|---|---|---|
| **1.** | Sample 1 | 800 | 40 | 0.1 | 65.8 |
| **2.** | Sample 2 | 800 | 50 | 0.2 | 65.78 |
| **3.** | Sample 3 | 800 | 60 | 0.3 | 67.4 |
| **4.** | Sample 4 | 1000 | 40 | 0.2 | 64.3 |
| **5.** | Sample 5 | 1000 | 50 | 0.3 | 69.9 |
| **6.** | Sample 6 | 1000 | 60 | 0.1 | 74.2 |
| **7.** | Sample 7 | 1200 | 40 | 0.3 | 58.3 |
| **8.** | Sample 8 | 1200 | 50 | 0.2 | 60.5 |
| **9.** | Sample 9 | 1200 | 60 | 0.1 | 64.6 |

To predict the hardness value at the nugget zone, two Machine Learning based algorithms, i.e., Random Forest and XG Boost, are used, and the results are compared with the results obtained from the Taguchi Algorithm. The process flowchart of implementing the Machine Learning algorithms on the experimental dataset is shown in Fig 4. The process parameters are optimized using the Taguchi $L_9$ orthogonal array methods. In this research, the aim is to maximize the hardness value of the target material. Therefore, a larger is better quality criterion was used to calculate the signal-to-noise ratio. RPM, Feed rate, and Plan depth were used as controllable process parameters.

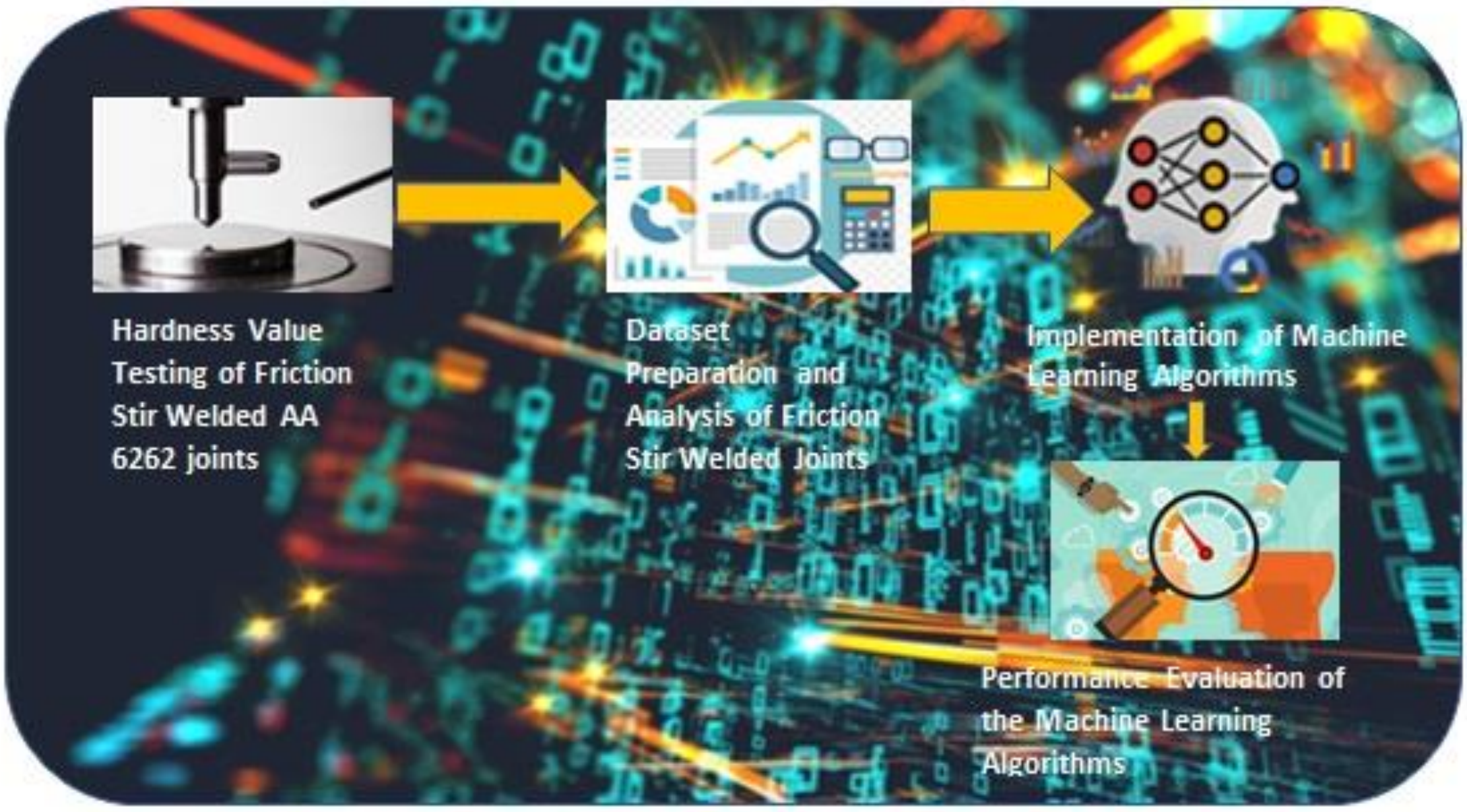

**Fig 4.** Process flowchart of the implementation of the Machine Learning algorithms on the Friction Stir Welded AA 6262 Hardness dataset.

## 3. Results and Discussion

### *3.1. Microstructure Property Analysis*

A transverse section of a friction stir weld can be clearly recognised due to the microstructure of the weld. The microstructure and characteristics of the material are altered during the FSW process. The degree of change is determined by the joint's location as well as a number of other factors, including the alloy type, starting microstructure, and the process and weld parameters employed to create the joint. The welding center is subjected to more heat and plastic deformation, resulting in a microstructural development that differs from the weld extremities, which are subjected to less heat and distortion. All of the process and weld factors are combined in the final microstructure. The microstructure of a material was investigated using a variety of microscopy techniques. Optical microscopy (OM), scanning electron microscopy (SEM), which offered topographic information, and Orientation Imaging Microscopy (OIM), which provided extra quantitative information, were the main techniques used. The linked plates and base materials were analyzed by evaluating optical microscope images. The large rupture surfaces were created using standard metallographic processes and etched with a reagent to analyze the grain structure of the welded zones and allow optical microscope analysis. Keller’s reagent was used to polish and etch cross-sections for optical microscopy, and OM was used for flushing, polishing, and studying the voids of the materials. OM observation depicted the changes in microstructure from the weld zone to the unaffected PM. The following considerations must be kept in mind during the specimen preparation process in order to expose a material's true microstructure. An appropriate etchant must be used to entirely erase the distortion. By sectioning, grinding, or polishing the specimen's surface, deformation can be implanted. The extent to which the effects of the deformation can be determined is determined by the degree of distortion. Gross deformation and plastic deformation are the two types of induced deformation. Figure 5 depicts the microstructures of nine samples.

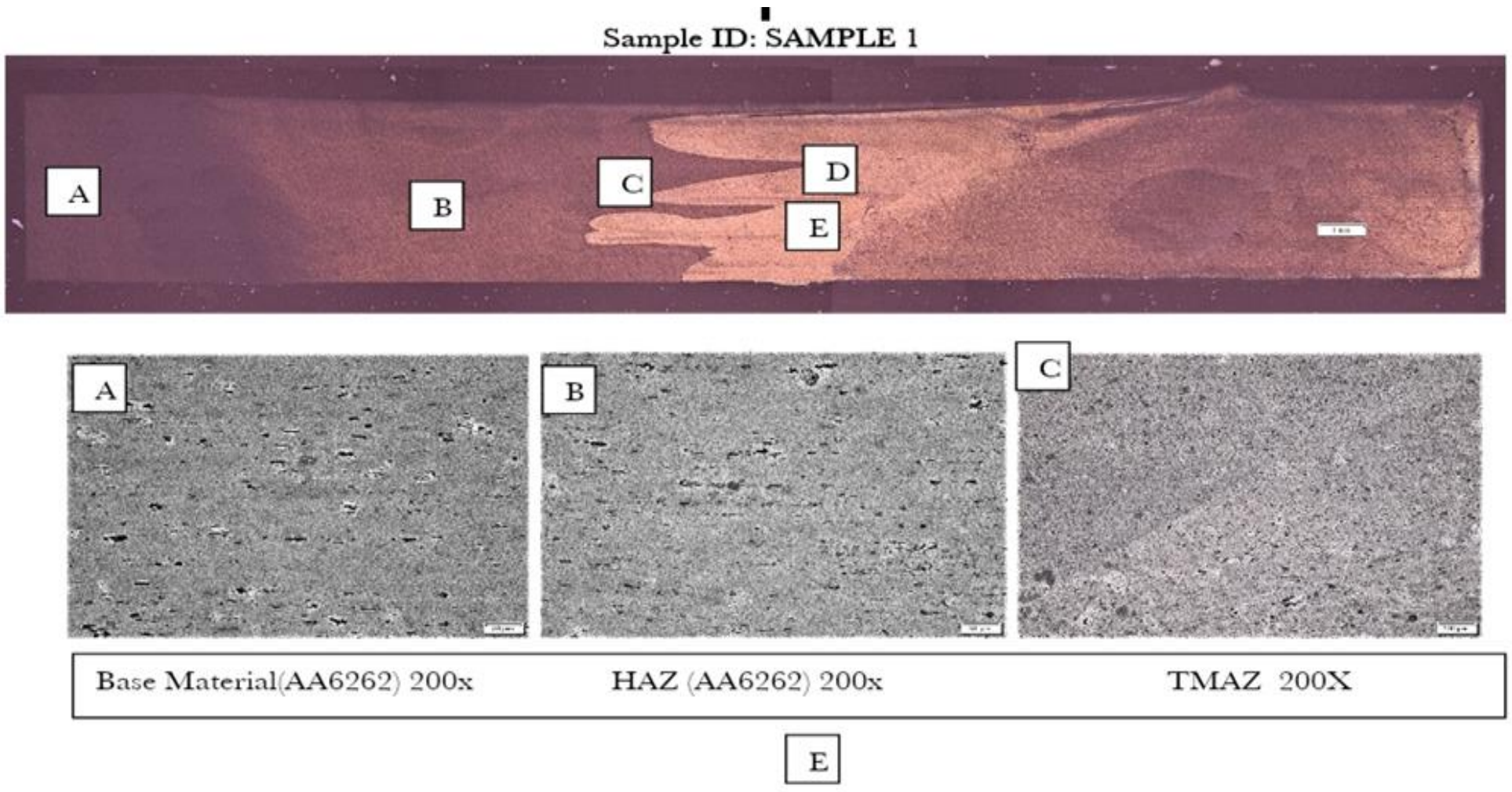

**Fig 5.** Microstructure obtained for Sample 1

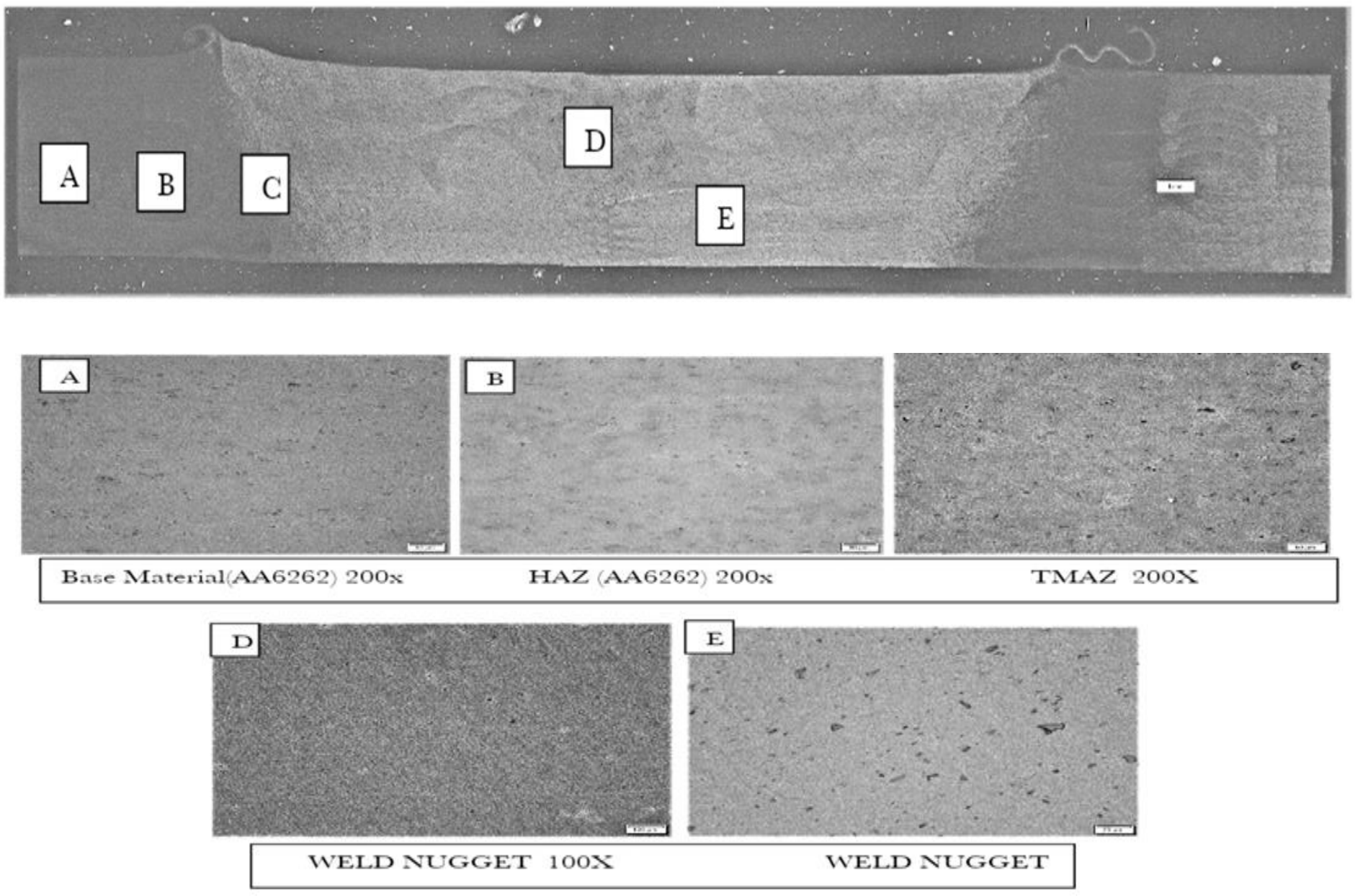

**Fig 6.** Microstructure obtained for Sample 2

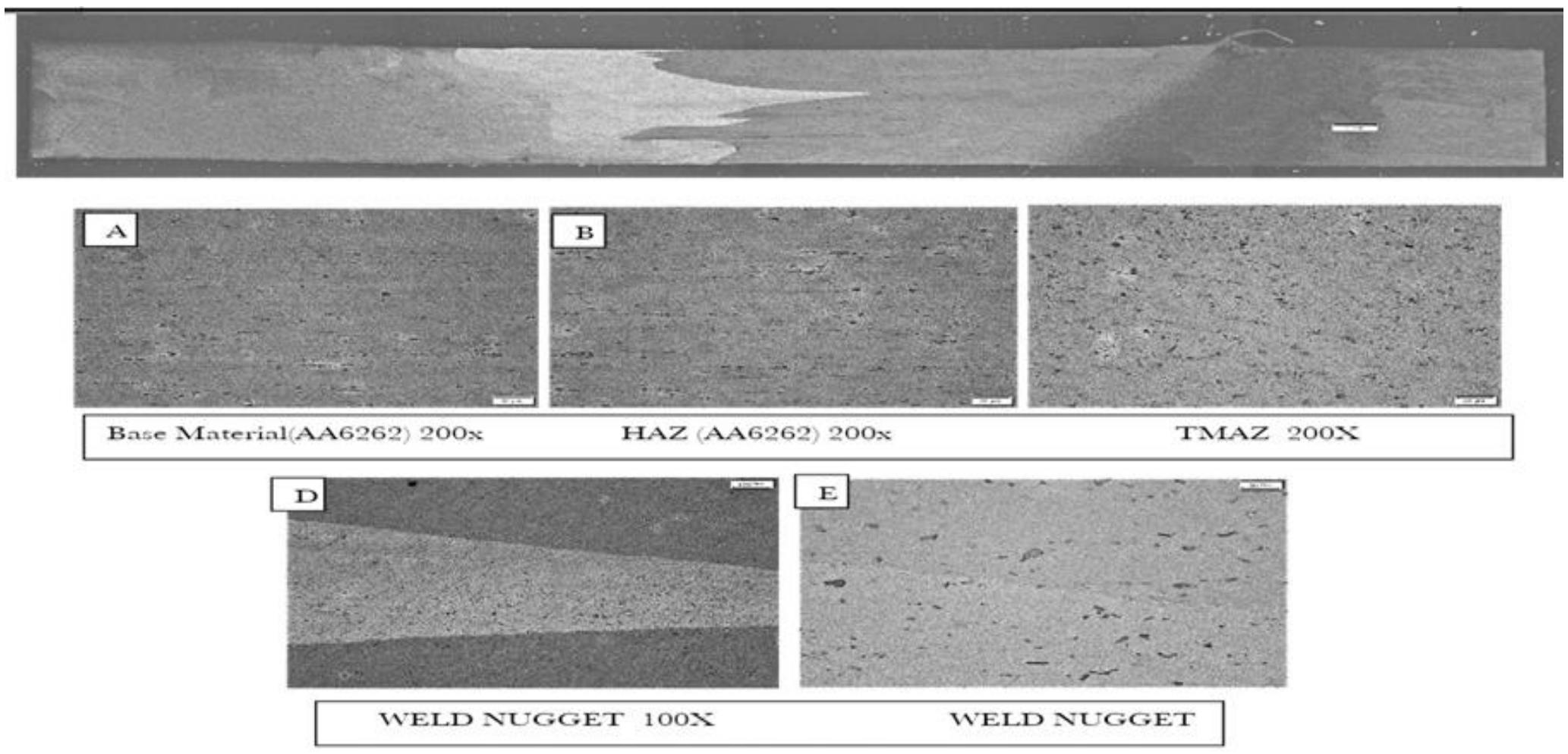

**Fig 7.** Microstructure obtained for Sample 3

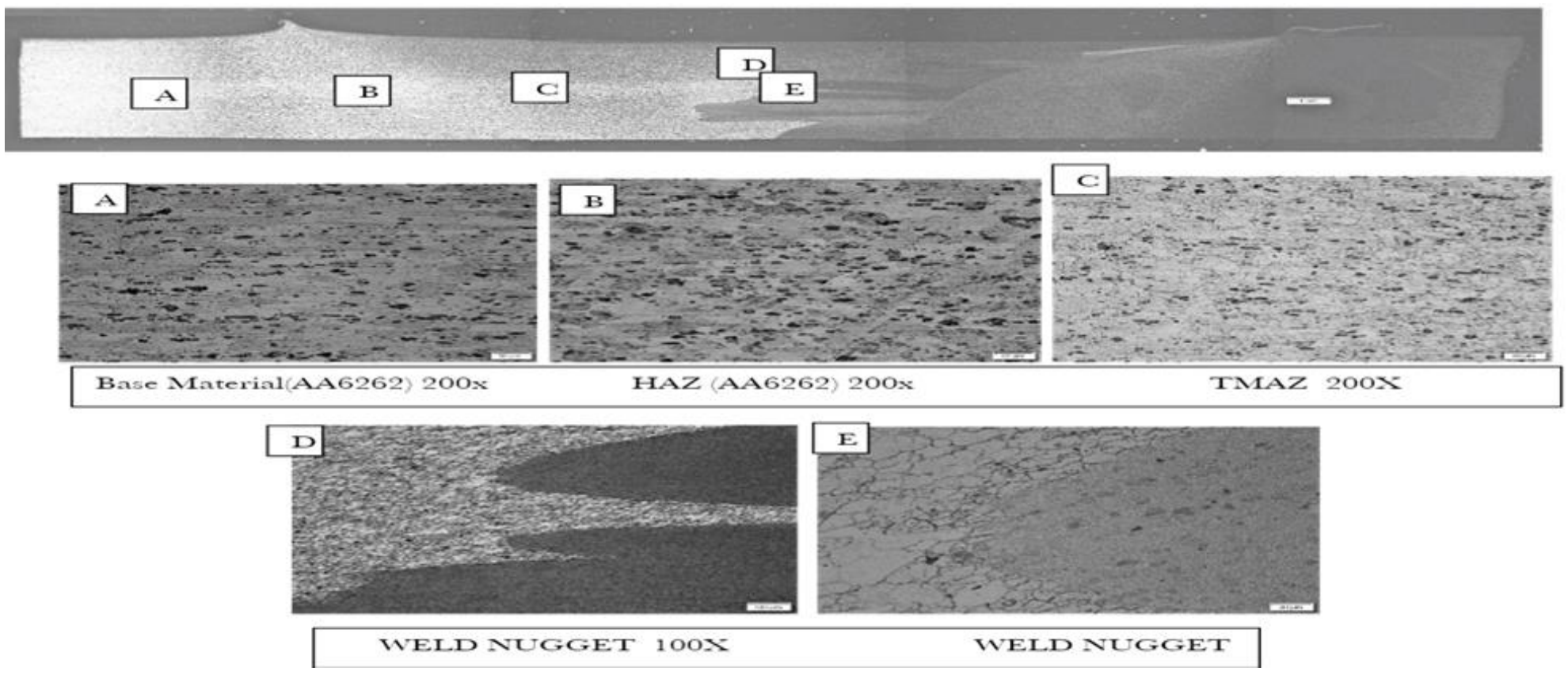

**Fig 8.** Microstructure obtained for Sample 4

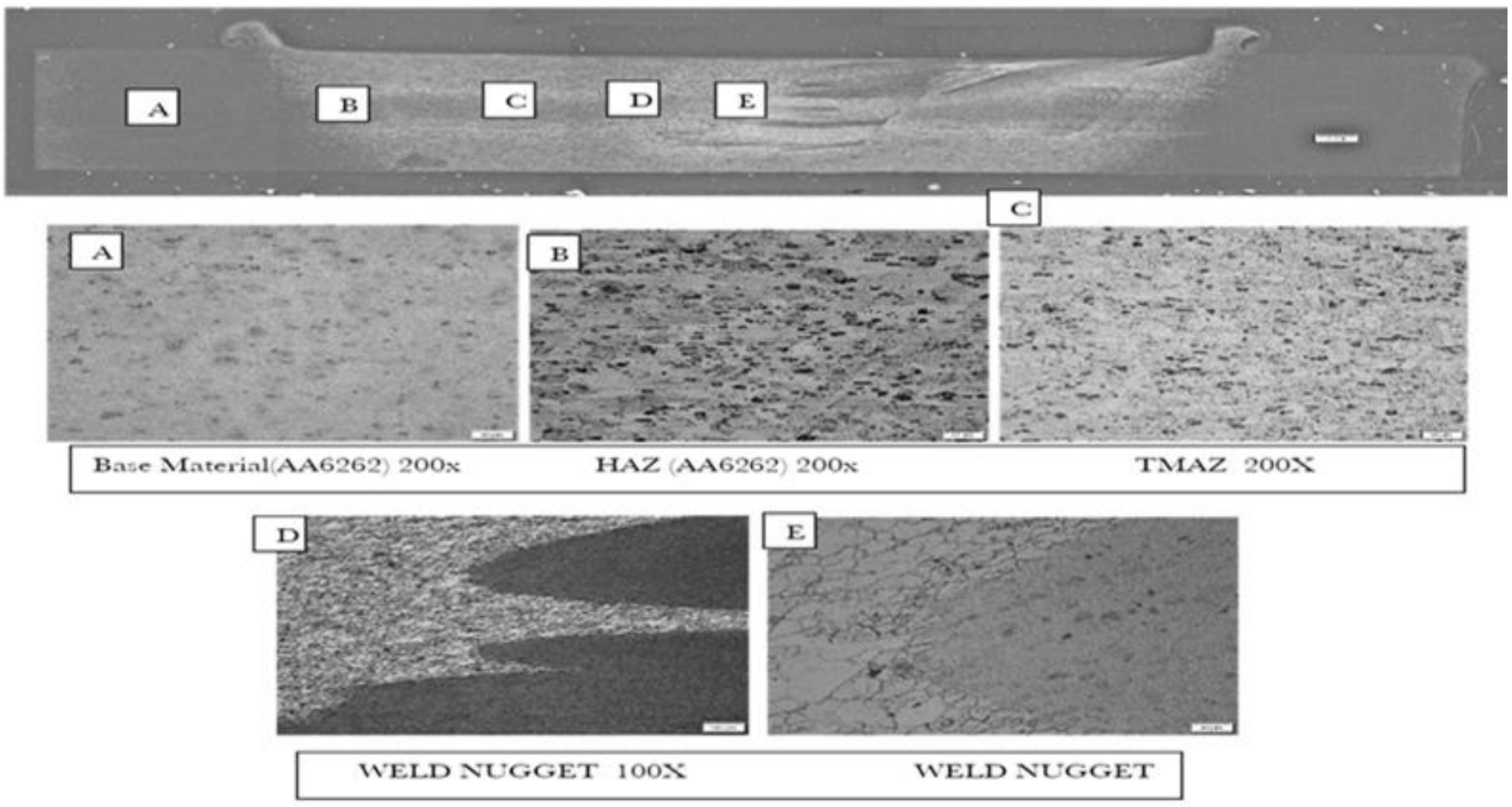

**Fig 9.** Microstructure obtained for Sample 5

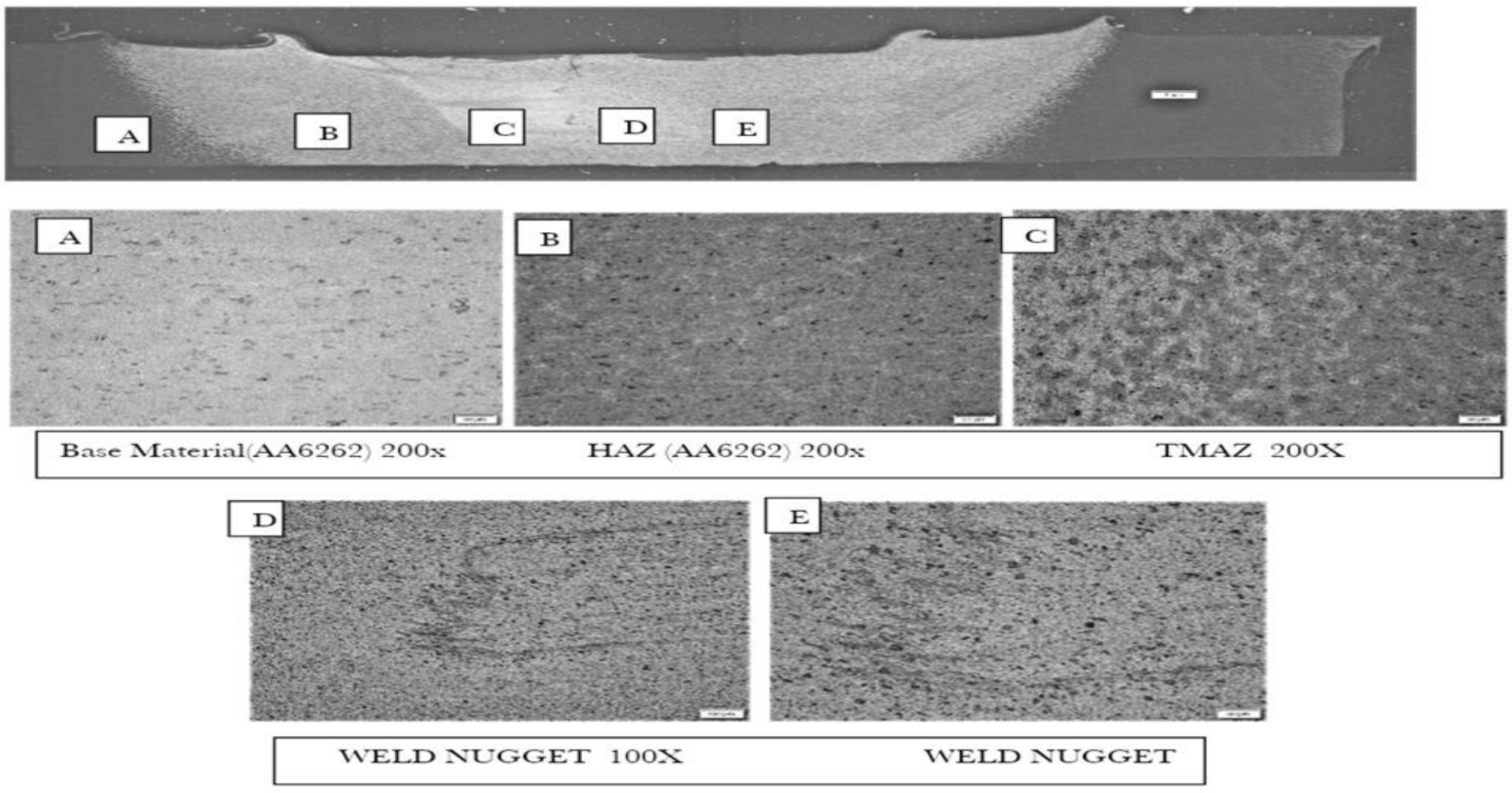

**Fig 10.** Microstructure obtained for Sample 6

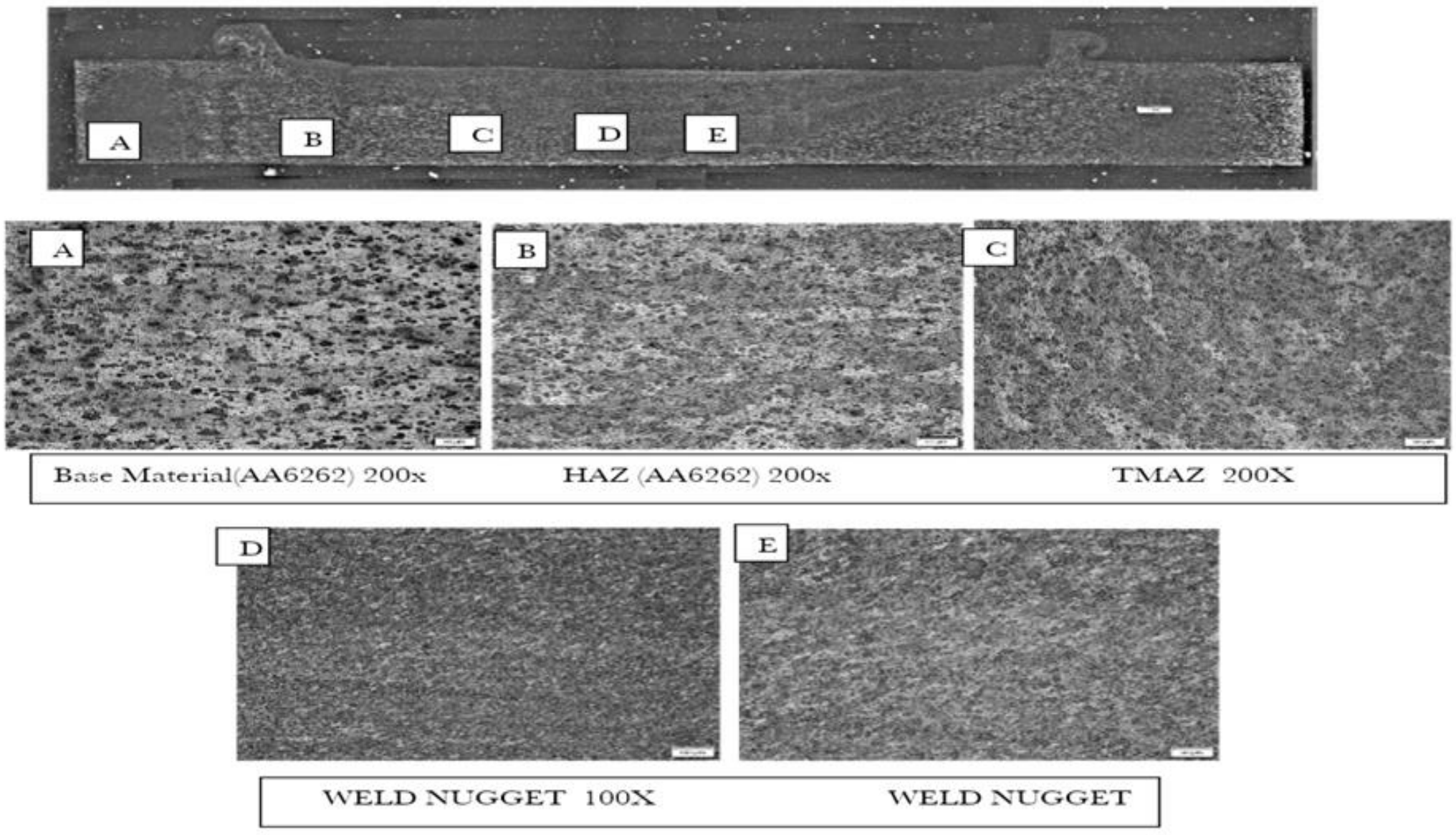

**Fig 11.** Microstructure obtained for Sample 7

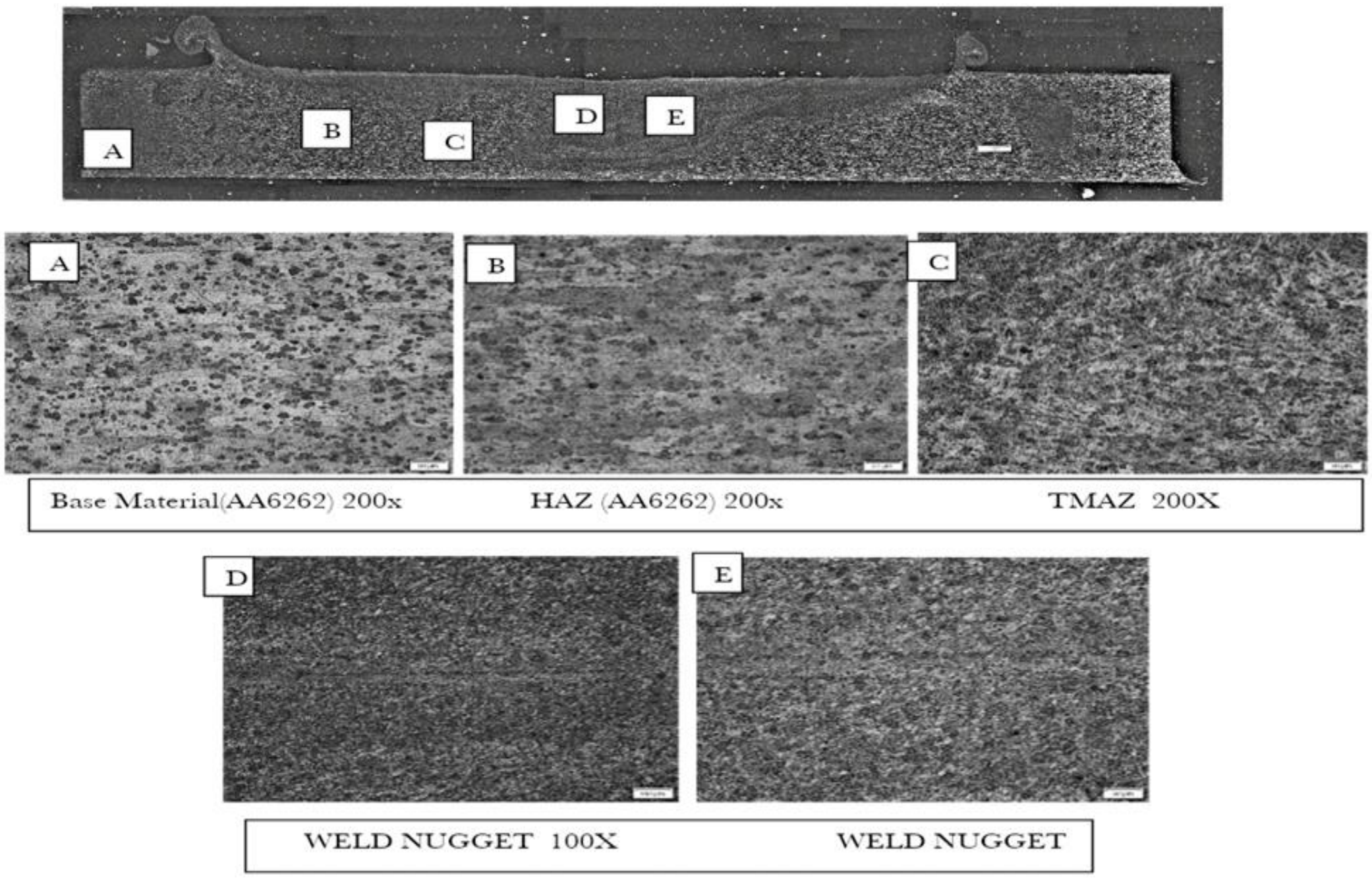

**Fig 12.** Microstructure obtained for Sample 8

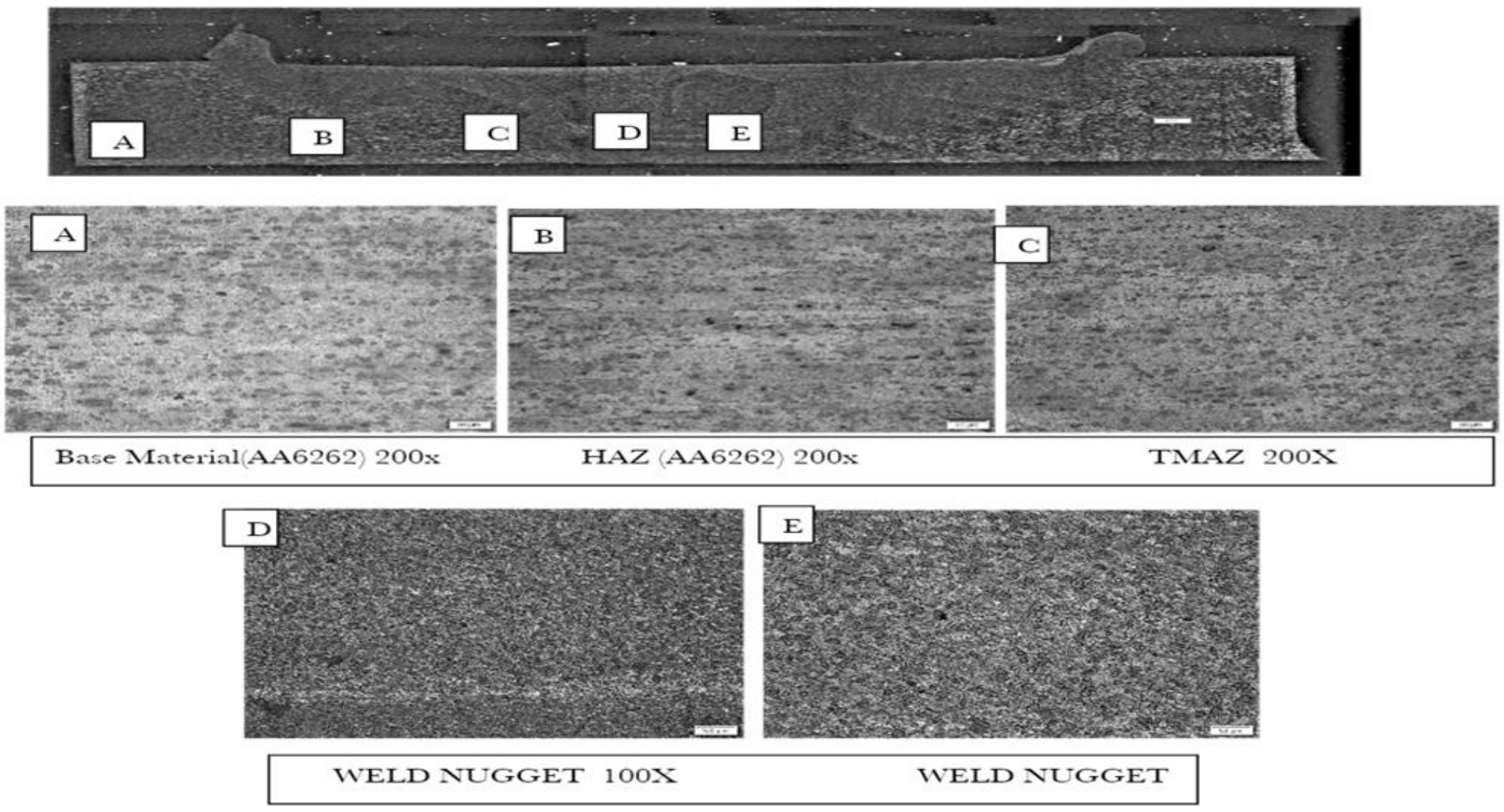

**Fig 13.** Microstructure obtained for Sample 9

### *3.2. Prediction of Hardness Value by Taguchi Algorithm*

The result indicated that in Fig 14, Table 3, and Table 4, Rotational speed and plan depth are the significant parameters for getting a higher hardness on the target materials. The maximum hardness of the weld joint at the nugget zone was observed at a combination of $A_3$, $B_2$, $C_3$, explicitly, a rotational speed of 1200 rpm, feed rate of 50 mm/sec, and plan depth of 0.3 mm imparts the optimum parameter combination. In addition, Analysis of Variance (ANOVA) was conducted to identify which parameter is significant or not. Moreover, the ANOVA result showed that rotational speed and plan depth are significant parameters for getting higher hardness to the target material at a 95% confidence interval.

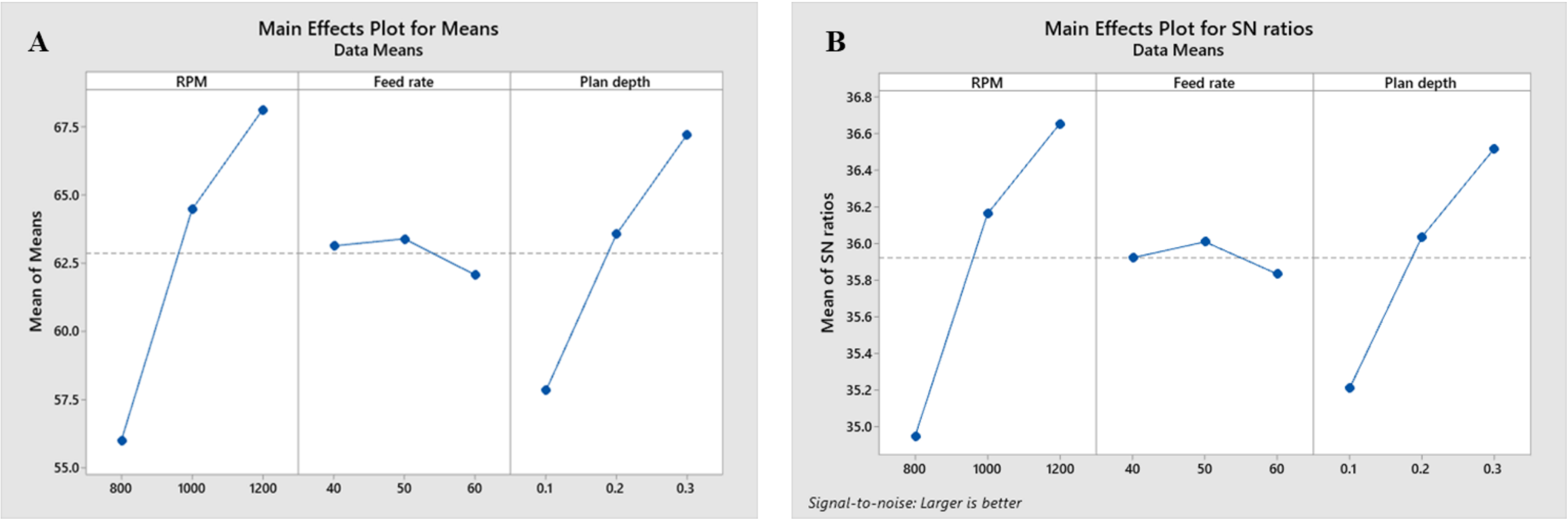

**Fig 14.** (a) Main effect means data plot, (b) S/N ratio data plot

**Table 3.** Analysis of Variance

| Source | DF | Adjusted SS | Adjusted MS | F-Value | P-Value |
|---|---|---|---|---|---|
| RPM | 2 | 232.621 | 116.311 | 145.62 | 0.007 |
| Feed rate | 2 | 2.965 | 1.483 | 1.86 | 0.350 |
| Plan depth | 2 | 133.779 | 66.889 | 83.75 | 0.012 |
| Error | 2 | 1.597 | 0.799 | | |
| Total | 8 | 370.963 | | | |

**Table 4.** Model Summary

| S | $R^2$ | $R^2$ (adjusted) | $R^2$ (predicted) |
|---|---|---|---|
| 0.89370 | 99.57% | 98.28% | 91.28% |

### *3.3. Hardness Value Prediction by Machine Learning Algorithms*

Exploratory Data Analysis (EDA) is a method for identifying patterns, anomalies, discrepancies, and other characteristics that best summarize a data set's main properties. This strategy entails employing various EDA approaches, including data visualization tools to gain insights into the data, validate the assumptions on which future inferences will be based, and even construct cautious models that define the data with the fewest possible variables. Fig 16 shows the exploratory data analysis result of the experimental dataset.

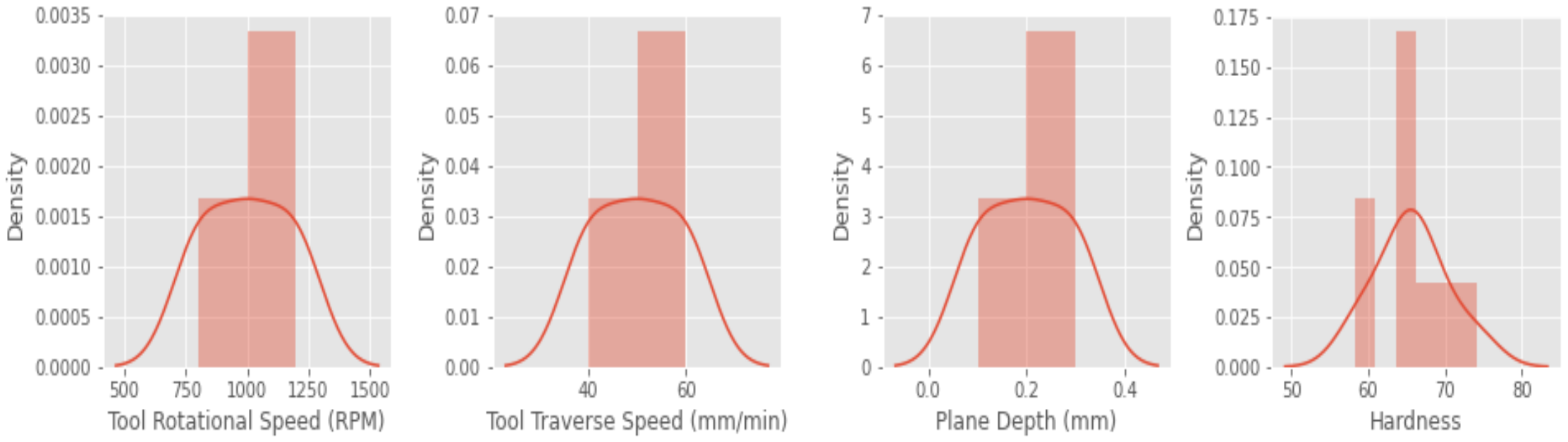

**Fig 15.** Exploratory Data Analysis of the experimental dataset

The process of picking the most significant features to input in machine learning algorithms is known as feature selection and is one of the primary components of feature engineering. Feature selection strategies reduce the number of input variables by removing redundant or unnecessary features and restricting the set of features to those most advantageous to the machine learning model. As the number and variety of datasets grow, it's more critical than ever to reduce them methodically. The fundamental purpose of feature selection is to improve the predictive model's performance while lowering the modeling cost. Fig 16 shows the feature importance plot of the input parameters from which it is observed that the Tool Rotational Speed (RPM) has more influence on the output parameter, i.e., hardness, which is further followed by the Tool Traverse Speed (mm/min).

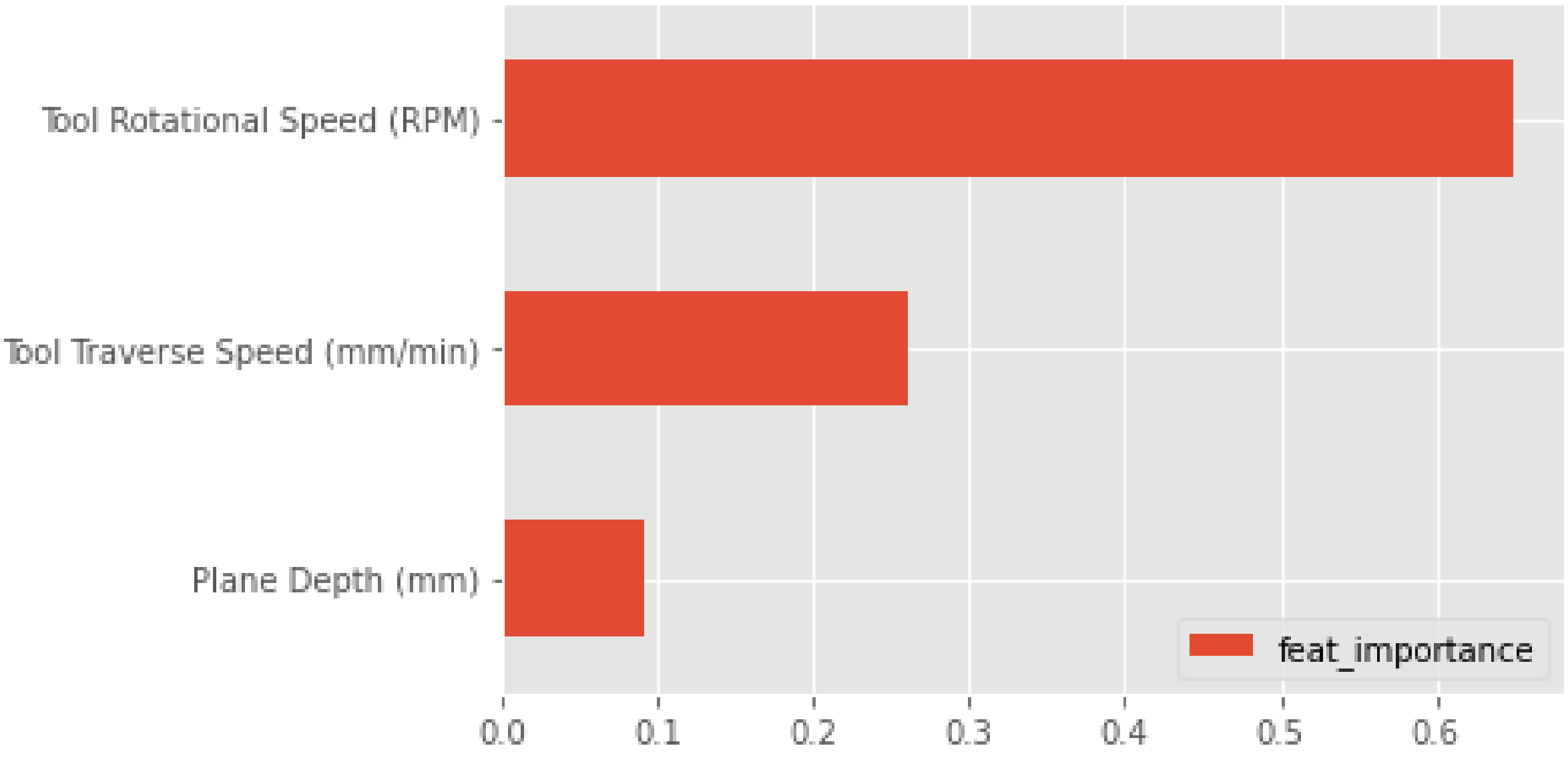

**Fig 16.** Calculation of the Feature Importance of input parameters

Ensemble learning is a type of machine learning that combines multiple base models to create a single best-in-class predictive model (powerful model). To consider a sample of Decision Trees, compute which characteristics to use or questions to ask at each split, and produce a final predictor based on the aggregated findings of the sampled Decision Trees, Ensemble Methods are used. Every sample is taken using replacement sampling. Each model is based on a different set of facts. The model is said to have high variance if it fluctuates a lot with changes in training data. Bagging is a strategy for minimizing model variance without influencing bias as a result. The random forest emerges from the decision tree's evolution to deliver more consistent outcomes. Many are preferable to one. To put it another way, the random forest method is based on this idea. In other words, a group of decision trees can give more accurate forecasts than a single decision tree. The random forest algorithm is a supervised classification strategy that makes more accurate and consistent predictions by combining N slightly different training decision trees. A decision tree is a flowchart-like structure with nodes and branches. The data is divided at each node based on one of the input properties, yielding two or more output branches. More and more splits occur in future nodes, resulting in an increasing number of branches that fragment the original data. This process is repeated until a node is generated in which all or almost all of the data belongs to the same class and there are no more splits or branches possible. The end outcome of this method is a tree-like structure. The first splitting node is the root node. The nodes at the end of the tree are called leaves, and they have a class designation. To determine which attribute leads to the purest subset, we must measure the purity of a dataset at each iteration. Various measurements and indices have been proposed in the past. The most commonly used are information gain, Gini index, and gain ratio. During training, the chosen quality measure is applied to all candidate features to evaluate which one generates the best split. The phrase entropy refers to the quantity of information or chaos in a system. It can also be used to determine the purity of a dataset. If the target classes are seen as possible statuses of a point in a dataset, the entropy of the dataset can be mathematically represented as the sum of all classes of the probability of each class multiplied by the logarithm of it. As a result, for a binary classification task, the entropy range is 0 to 1. Equation 1 is used to calculate entropy (p).

$$Entropy\ (p) = -\sum_{i=1}^{N} p_i log_2 p_i \qquad (1)$$

Where p is the dataset's total size, N denotes the number of classes, and pi denotes the frequency of class I within the dataset. The difference in entropy before and after the split is calculated to determine how good a feature is for splitting. To put it another way, calculate the dataset's entropy before splitting it into subgroups, and then compute the entropy of each subset after splitting. Finally, the sum of the output entropies is subtracted from the dataset's entropy, which is weighted by the size of the subgroups, before the split. This discrepancy indicates whether information has risen or reduced in entropy.

$$Informtion\ Gain = Entropy(before) - \sum_{j=1}^{K} Entropy(j, after) \qquad (2)$$

Where "before" is the dataset before the split, "K" denotes the number of subsets created by the split, and (j, after) denotes subset j after the split. We'd then split the data at each phase based on the feature with the highest information gain value, as this produces the purest subsets. This statistic is used by the ID3 algorithm. The ID3 technique has the disadvantage of favoring features with a larger number of values, which results in longer decision trees. The SplitInfo concept is introduced by the C4.5 algorithm's gain ratio metric. SplitInfo is the sum of the weights multiplied by the logarithm of the weights. The weights are the ratio of the number of data points in the current subset to the number of data points in the original dataset. The gain ratio is derived by dividing the ID3 algorithm's information gain by the SplitInfo value, as shown in Equation 3.

$$Gain\ Ratio = \frac{Information\ Gain}{SpitInfo} = \frac{Entropy(before) - \sum_{j=1}^{K} Entropy(j, after)}{\sum_{j=1}^{K} w_j log_2 w_j} \qquad (3)$$

Where "before" is the dataset before the split, "K" denotes the number of subsets created by the split, and (j, after) denotes subset j after the split. The CART algorithm also uses the Gini index to determine purity or impurity. The Gini index is based on Gini impurity. As seen in equation 4, Gini impurity is defined as 1 minus the sum of the squares of the class probabilities in a dataset.

$Gini\ Impurity\ (p) = 1 - \sum_{i=1}^{N} p_i{}^2$ (4)

Table 5 shows the result obtained by implementing the Random Forest algorithm on the experimental dataset.

**Table 5:** Metric Features for evaluating the performance of Random Forest Algorithm

| Mean Square Error (MSE) | Mean Absolute Error | $R^2$ |
|---|---|---|
| 4.42 | 1.979 | 0.62 |

The decision tree obtained for the experimental dataset is shown in Fig 17.

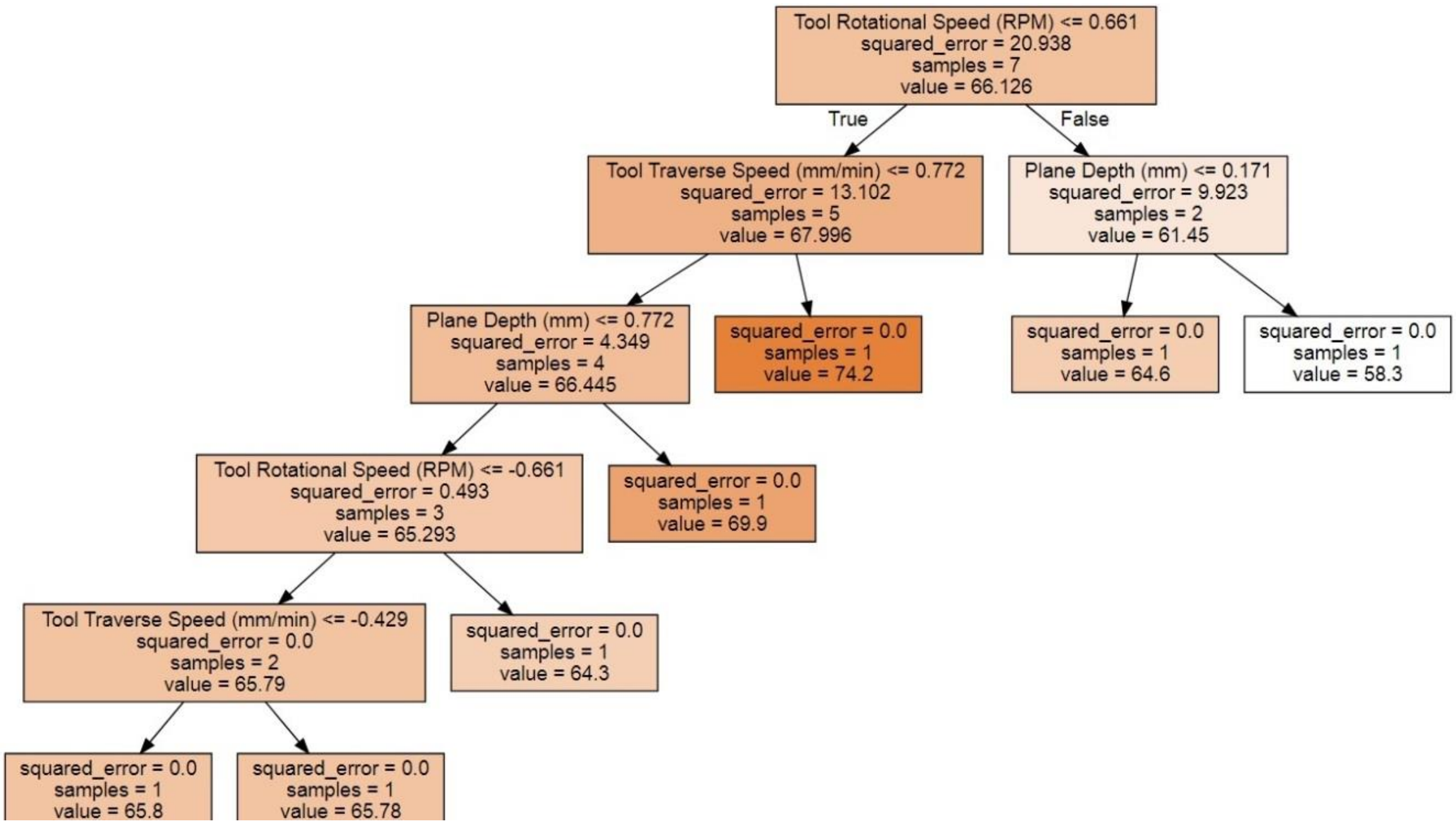

**Fig 17.** Decision Tree obtained on the experimental dataset

XGBoost is a method of ensemble learning. It's not always enough to rely solely on the output of a single machine learning model. Ensemble learning is a way for progressively combining several trainees' prediction skills. As a result, a single model is developed that combines the results of several models. The foundation learners, or models, for the ensemble could come from the same or distinct learning algorithms. Ensemble learners are divided into two types: bagging and boosting. Though these two tactics can be used with various statistical models, decision trees have proven to become the most popular. Three simple steps make up the boosting ensemble technique:

1. To predict the target variable of y, an initial model F0 is defined. A residual (y – F0) will be associated with this model.
2. The residuals from the previous step are fitted to a new model, h1.
3. F0 and h1 are now joined to produce F1, an enhanced version of F0. The F1 means the squared error will be lower than the F0 mean squared error, as shown in equation 5.

$F_1(x) < F_0(x) + h_1(x)$ (5)

To increase the performance of F1, we may develop a new model, F2, based on F1's residuals, as shown in equation 6.

$F_2(x) < F_1(x) + h_2(x)$ (6)

This can be repeated for 'm' iterations until the residuals are as low as possible, as shown in equation 7.

$F_m(x) < F_{m-1}(x) + h_m(x)$ (7)

The additive learners do not interfere with the functions generated in the previous phases in this step. Instead, they provide their information to reduce inaccuracies.

Table 5 shows the obtained result on the experimental dataset.

**Table 5.** Metric Features for evaluating the performance of XGBoost Algorithm

| Mean Square Error (MSE) | Mean Absolute Error | $R^2$ |
|---|---|---|
| 4.14 | 1.99 | 0.65 |

## 4. Conclusion

In the present study, three algorithms (one Taguchi method-based and two machines learning-based) were subjected to the Hardness value of the Friction Stir Welded AA6262 joints. From the implementation of the Machine Learning algorithms, the following conclusions are made:

- The $R^2$ value of the XGBoost algorithm is more than the Random Forest algorithm, which means it gives more accurate results in comparison to Random Forest.
- Increasing makes use of trees with fewer splits than bagging systems like Random Forest, which use trees that have been fully formed. The interpretation of such little trees, which aren't very deep, is straightforward.
- The number of trees or iterations, the rate at which the gradient boosting learns, and the depth of the tree can all be modified using validation techniques such as k-fold cross-validation. Overfitting may occur if there are a large number of trees present. As a result, the boosting stopping circumstances must be selected with care. From the implementation of the Taguchi algorithm, the following conclusions are drawn:

- Higher temperatures are developed at the nugget zone when the rotational speed and plane depth increase. Therefore, the temperature is one of the significant physical factors for getting sound welds. In addition to this, the maximum hardness value was observed at higher rotational speed and plan depth.

- The rotational speed of 1200 rpm and plan depth of 0.3 mm has bestowed the maximum hardness values at the nugget zone. The ANOVA result showed that rotational speed and plan depth has the most significant parameters for getting higher hardness of the target material.

- The coefficient of the determination, i.e., the $R^2$ value, is 0.91, which is higher than the implemented Machine Learning algorithms.

## Declaration

On behalf of all authors, the corresponding author states that there is no conflict of interest.